\documentclass[letterpaper]{article} 
\usepackage{aaai25}  
\usepackage{xcolor}
\usepackage{times}  
\usepackage{helvet}  
\usepackage{courier}  
\usepackage[hyphens]{url}  
\usepackage{graphicx} 
\usepackage{amsmath} 
\usepackage{amssymb} 
\usepackage{multirow}
\usepackage{multicol}
\usepackage{makecell}
\usepackage{array}
\usepackage{arydshln}
\usepackage{subfigure}
\usepackage{natbib}  
\usepackage{caption} 
\usepackage{algorithm}
\usepackage{algorithmic}

\usepackage{newfloat}
\usepackage{listings}
\DeclareCaptionStyle{ruled}{labelfont=normalfont,labelsep=colon,strut=off} 
\floatstyle{ruled}
\newfloat{listing}{tb}{lst}{}
\floatname{listing}{Listing}
\nocopyright 

\title{SentenceVAE: Enable Next-sentence Prediction for Large Language Models with Faster Speed, Higher Accuracy and Longer Context}
\author{
    \
    Hongjun An\textsuperscript{\rm 1,}\textsuperscript{\rm 2}\equalcontrib,
    Yifan Chen\textsuperscript{\rm 1,}\textsuperscript{\rm 2}\equalcontrib,
    Zhe Sun\textsuperscript{\rm 1,}\textsuperscript{\rm 2}\thanks{Corresponding author: sunzhe@nwpu.edu.cn} \& 
    Xuelong Li\textsuperscript{\rm 1,}\textsuperscript{\rm 2}\thanks{Corresponding author: li@nwpu.edu.cn}
}

\affiliations{
    \textsuperscript{\rm 1}School of Artificial Intelligence, OPtics and ElectroNics (iOPEN), Northwestern Polytechnical University\\

    \textsuperscript{\rm 2}Institute of Artificial Intelligence (TeleAI), China Telecom
    
}

\usepackage{bibentry}

\begin{document}

\maketitle

\begin{abstract}
Current large language models (LLMs) primarily utilize next-token prediction method for inference, which significantly impedes their processing speed. In this paper, we introduce a novel inference methodology termed next-sentence prediction, aiming at enhancing the inference efficiency of LLMs. We present Sentence Variational Autoencoder (SentenceVAE), which includes a Sentence Encoder to compress multiple tokens in a sentence into a single token, and a Sentence Decoder to reconstruct it. By integrating SentenceVAE into the input and output layers of LLMs, we develop Sentence-level LLMs (SLLMs) that employ a sentence-by-sentence inference method. In addition, the SentenceVAE module of SLLMs can maintain the integrity of the original semantic content by segmenting the context into sentences, thereby improving accuracy while boosting inference speed. Moreover, compared to previous LLMs, SLLMs process fewer tokens over equivalent context length, significantly reducing memory demands for self-attention computation and facilitating the handling of longer context. Extensive experiments on Wanjuan dataset have revealed that the proposed method can accelerate inference speed by 204$\sim$365\%, reduce perplexity (PPL) to 46$\sim$75\% of its original metric, and decrease memory overhead by 86$\sim$91\% for the equivalent context length, compared to previous token-by-token methods.
\end{abstract}

%

\section{Introduction}

\begin{figure}[!t]
  \centering
  \includegraphics[width=0.9\columnwidth]{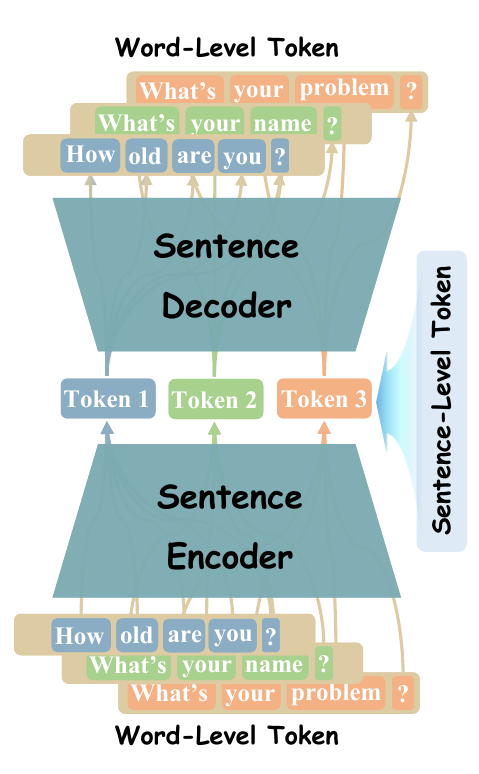}
  \caption{The schematic form of SentenceVAE. It clearly illustrates that the encoder of SentenceVAE can compress the information contained within a sentence into a single token, and the decoder can restore the compressed token back to its original sentence form.}
  \label{sample}
\end{figure}

Large language models (LLMs) have demonstrated remarkable proficiency in understanding human intent through the acquisition of knowledge from vast and diverse datasets. These models provide rich and accurate responses, making them essential for applications such as machine translation \cite{pmlr-v202-zhang23m, wang-etal-2023-document-level}, chatbots \cite{islam2024gpt, wang2024telechat}, and question-answering systems \cite{singhal2023towards, lazaridou2022internet}. The predominant method to inference in LLMs involves next-token prediction, where the model generates tokens sequentially. While this token-by-token generation method has achieved significant performance, it is inherently limited in producing only one token per inference step. This constraint introduces considerable computational overhead, prolonging the inference process and impeding the scalability and operational efficiency of LLMs. Consequently, there is a critical need to explore alternative methods that can improve the inference speed while preserving or even enhancing the accuracy and responsiveness of LLMs.

To enhance the inference speed of LLMs, Gloeckle et al. \cite{gloeckle2024better} proposed training LLMs to predict multiple tokens simultaneously, aiming to accelerate the inference process. However, the number of tokens predicted simultaneously is fixed, resulting in a rigid and inflexible partitioning of the input context. For example, when set to five, the input is invariably segmented into non-overlapping sequences of five tokens, regardless of the context's underlying structure, potentially compromising inference accuracy.

To enhance inference speed while preserving or even enhancing accuracy, we explore method for adaptively selecting the optimal number of predicted tokens based on the semantic content of the text. Consequently, we introduce Sentence Variational Autoencoder (SentenceVAE), which comprises a sentence encoder and a sentence decoder. The schematic of SentenceVAE is illustrated in Fig. \ref{sample}. The sentence encoder is designed to condense the information of an entire sentence into a single token, while the sentence decoder reconstructs this compressed token back into a sentence. By seamlessly incorporating the encoder and decoder components of SentenceVAE into the input and output layers of LLMs, these enhanced models are capable of performing next-sentence prediction. This integration not only enriches the models' capabilities but also accelerates the inference speed, leading to improved efficiency in language understanding and generation. Moreover, since SentenceVAE segments text at the sentence level, the semantic integrity is better preserved, ensuring the accuracy of the inferrence. Specifically, prior to inputting text into LLMs, SentenceVAE segments the text into sentences and utilizes the sentence encoder to compress the information contained within each sentence into a single token. These compressed tokens are then fed as input to the LLMs. The LLMs predict their output based on these compressed tokens. Finally, the sentence decoder decodes the predicted token into the final output of the LLMs.

The contribution of the proposed method can be described as follows:

\begin{itemize}
\item We propose SentenceVAE, a model capable of compressing the information content of a sentence into a single token and expanding such tokens to reconstruct sentences with high informational fidelity.
\item We embed the encoder and decoder components of SentenceVAE into the input and output layers of LLMs, enabling these models to implement an effective inference method for next-sentence prediction.
\item Extensive experiments conducted on various LLMs have demonstrated that the proposed method not only accelerates inference speed, but also improves accuracy, and reduces memory usage, enabling the model to handle longer contexts with the same device memory.
\end{itemize}

\begin{figure*}
  \centering
  \includegraphics[width=\textwidth]{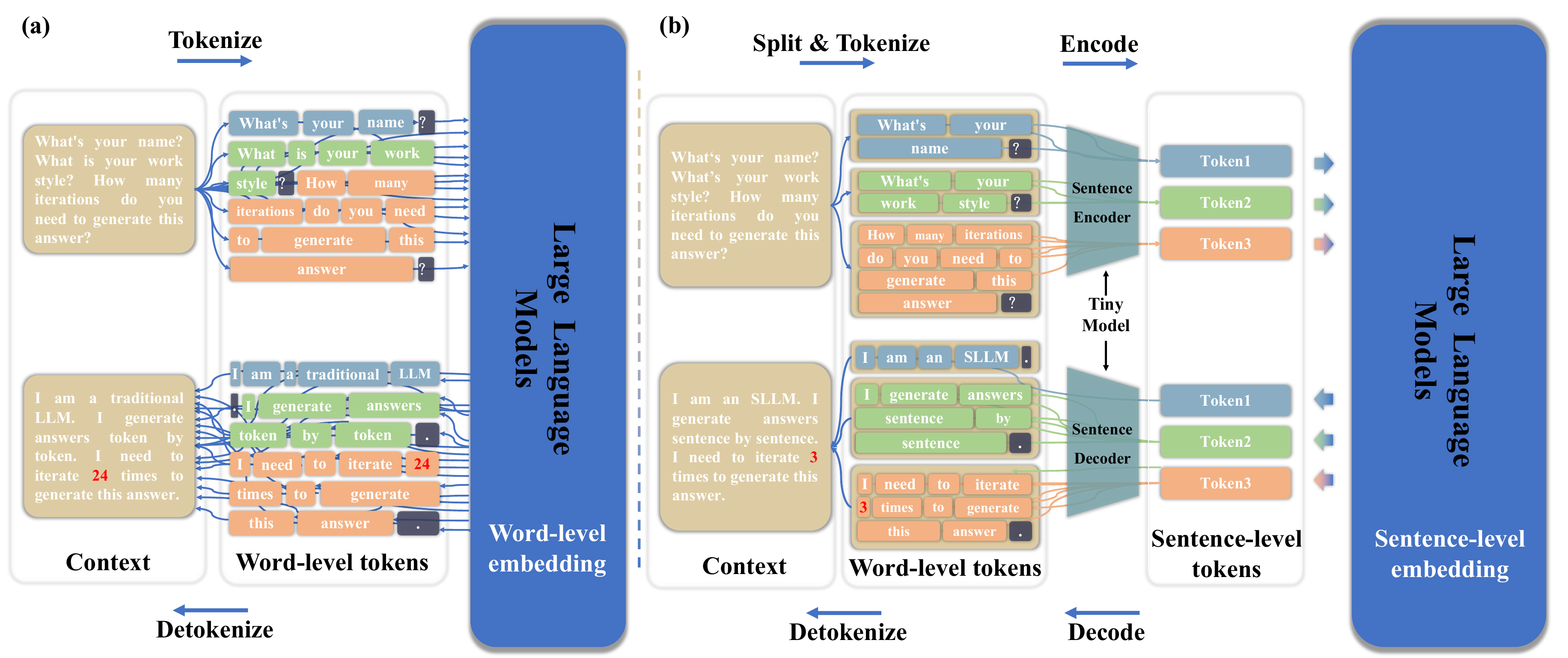}
  \caption{(a) The schematic form of published LLMs. (b) The schematic form of SLLMs, which embedded with SentenceVAEs. It can be clearly seen that, unlike the next-token inference method employed by published LLMs, the proposed method adopts a next-sentence prediction method, which significantly reduces the number of inference iterations and the overall inference cost. }
  \label{frame}
\end{figure*}

\section{Related Work}

{\bf Large language models:} recently, LLMs have achieved remarkable performance in the field of natural language processing (NLP). However, these models typically possess a vast number of parameters from billions to hundreds of billions, exemplified by models such as TeleChat (1B-52B) from TeleAI \cite{wang2024telechat, li2024teleflm}, Llama (7B-405B) from Meta \cite{touvron2023llama, touvron2023llama2}, InternLM (1.8B-20B) from Shanghai AI Lab \cite{cai2024internlm2}, etc. The sheer magnitude of these parameters renders the inference process computationally intensive and time-consuming. Moreover, most LLMs employ a one-token prediction method, where the model predicts only one token at a time. This token-by-token method further exacerbates the inference time of LLMs. As a result, accelerating the inference speed of LLMs while preserving or enhancing their accuracy presents a significant challenge for researchers.

{\bf Multi token prediction:} the transformer model enables the parallel processing of sequences through its self-attention mechanism, providing a technical foundation for the implementation of multi-token prediction \cite{vaswani2017attention}. The BERT model utilizes bidirectional encoder representations to capture contextual information, further solidifying the basis for the implementation of multi-token prediction \cite{devlin2018bert}. The T5 model incorporates multi-token prediction during the training process for the first time, significantly enhancing the coherence and quality of the generated text \cite{raffel2020exploring}. The Better \& Faster LLMs apply multi-token prediction to LLMs, allowing them to directly predict multiple tokens in a single inference step, thereby reducing the number of inference iterations and associated computational costs \cite{gloeckle2024better}. However, the number of predicted tokens is fixed, which may inadvertently combine unrelated tokens. This can result in the merging of unrelated tokens, disrupting the original sentence structure and consequently reducing inference accuracy.

{\bf Encoder-decoder model:} the concept of an encoder is introduced by Yann LeCun in his doctoral dissertation \cite{lecun1987}, wherein he posited that an encoder could transform input data into shorter representations without losing the core information, thereby providing a foundational approach to compressing high-dimensional data. The Variational Autoencoder (VAE) model introduces the use of hidden variables between the encoder and decoder to learn the data distribution, laying the groundwork for deep-learning-based encoders and decoders \cite{kingma2013auto}. Sutskever et al. introduces the encoder-decoder framework into the field of deep learning, utilizing recurrent neural networks (RNNs) as both the encoder and decoder for sequence-to-sequence (Seq2Seq) learning, demonstrating exceptional performance in machine translation tasks \cite{sutskever2014sequence}.
The SegNet applies the encoder-decoder model to the field of image segmentation, significantly enhancing segmentation performance \cite{badrinarayanan2017segnet}. Subsequently, the encoder-decoder model has achieved great success in various domains including machine translation \cite{cho2014properties, makin2020machine}, text-to-speech conversion \cite{li2020robutrans}, computational imaging \cite{chen2023computational, li2024part}, and beyond, fully demonstrating the capability of the encoder-decoder architecture to efficiently compress and restore data.

By leveraging the advantages of multi-token prediction and the encoder-decoder model, we propose SentenceVAE. SentenceVAE is capable of compressing the information contained within a sentence into a single token, enabling next-sentence prediction and significantly accelerating the inference speed of LLMs.  Additionally, our method divides the context into sentences without compromising the original semantics, accelerating the inference speed of LLMs while ensuring inference accuracy.

\section{Method}

We revisited the mechanisms underlying human interaction. Typically, when individuals prepare to speak, they have already formulated the semantic content of an entire sentence in their minds. However, due to the limitations of the vocal cords, which are relatively inefficient organs for sound production, speakers are compelled to articulate sentences word-by-word, much like the token-by-token generation method of LLMs. Similarly, listeners tend to comprehend sentences holistically, rather than word-by-word. This suggests that sentences are the fundamental units of comprehension in natural language. Inspired by this, we developed the Sentence Variational Autoencoder (SentenceVAE), which includes a Sentence Encoder to compress a sentence into a single token and a Sentence Decoder to reconstruct it. By integrating SentenceVAE into LLMs, these models can operate within a more efficient sentence-level embedding space, enabling next-sentence prediction. This method significantly reduces the number of tokens for the same context length, thereby improving the model's inference speed and reducing the memory usage required for self-attention calculation, enabling the model to handle longer contexts. Additionally, by understanding language at the sentence level, the model demonstrates improved accuracy.


\subsection{Sentence Variational Autoencoder (SentenceVAE)}

Our SentenceVAE primarily consists of a Sentence Encoder and a Sentence Decoder. The encoder encodes multiple word-level tokens from a sentence into a single sentence-level token, while the decoder reconstructs this sentence-level token back into the original sequence of word-level tokens. To ensure that the multiple tokens input to the encoder come from individual sentences, we propose a specialized sentence segmentation mechanism. Additionally, we introduce a feature fusion mechanism, which enable the encoder to encode variable-length token sequences into a single sentence-level token.

{\bf Sentence Segmentation Mechanism:} to ensure that all word-level tokens input to the sentence encoder per time come from a single sentence, we use regular matching to split sentences by punctuation marks such as \texttt{","}, \texttt{"."}, \texttt{"?"}, \texttt{"!"} before tokenizing. Let the original string be $S$, and after partitioning, the set $\{s_i|1\leq i \leq n\}$ is obtained, satisfying $S=s_1 s_2 s_3 ... s_n$.

{\bf Sentence Encoder:} taking $s = s_t (1\leq t \leq n)$ as an example, after tokenization, we obtaine a sequence of $L$ token ids $\boldsymbol{D}=[d_1, d_2, d_3, ..., d_L]$ representing a sentence. This sequence is then passed through an embedding layer, resulting in word-level embedding vectors of $\boldsymbol{E}$, as shown in Eq.\ref{eq_senc_emb}. 

\begin{equation}
    \label{eq_senc_emb}
    \boldsymbol{E} = (\boldsymbol{e}_i)_{L\times \text{hidden\_size}} = \textbf{Embed}(\boldsymbol{D}) 
\end{equation}

These embeddings are subsequently input into self-attention based encoder blocks, yielding hidden features $\boldsymbol{H}$, as shown in Eq. \ref{eq_senc_hidden}. 

\begin{equation}
    \label{eq_senc_hidden}
    \boldsymbol{H} = (\boldsymbol{h}_i)_{L\times \text{hidden\_size}} = \textbf{EncoderBlocks}(\boldsymbol{E})
\end{equation}

To derive the sentence-level token, we fuse these $L$ features into a single vector $\boldsymbol{\Omega}_t \in \mathbb{R}^{\text{hidden\_size}}$. This sentence-level token is then fed into the decoder-only LLM, which generates a new sentence-level token $\boldsymbol{\Omega}_{t+1}$ at each time step.

{\bf Feature Fusion Mechanism:} after the encoder blocks generates $\boldsymbol{H}$, we propose a method to fuse them into a single sentence-level token $\boldsymbol{\Omega}_t$. This method involves accumulating the $L$ vectors and normalizing them using Layer Normalization \cite{ba2016layer}, as Eq. \ref{eq_ffm_da}. 

\begin{equation}
    \label{eq_ffm_da}
    \boldsymbol{\Omega}_t = \textbf{LayerNorm}(\sum_{i=1}^L{\boldsymbol{h}_i}) 
\end{equation}

{\bf Sentence Decoder:} the Sentence Decoder contains a combination of masked self-attention and cross-attention blocks, where the cross-attention mechanism uses the individual sentence embedding vector $\boldsymbol{\Omega}_{t+1}$ as key (K) and value (V).

In the prediction phase, given $\boldsymbol{\Omega}_{t+1}$, initialize the input token $d_0=<bos>$, the decoder outputs $d_1$, and then input $[d_0, d_1]$ and $\boldsymbol{\Omega}_{t+1}$ into the decoder. Repeat this process until $d_{l'+1}=<eos>$, and terminate the iteration, as Eq. 
 {\ref{eq_sdec_pred}}.

\begin{equation}
    \label{eq_sdec_pred}
    \begin{split}
        \textbf{init: } & d_0= <bos>, l' = 0 \\
        \textbf{do: } & \\
        & \boldsymbol{D}_{l'} = [d_0, d_1, d_2, ..., d_{l'}] \\
        & \boldsymbol{P}_{l'} = (\boldsymbol{p}_i)_{(l'+1) \times \text{hidden\_size}} \\
        &= \textbf{DecoderBlocks}(\boldsymbol{D}_{l'}, \boldsymbol{\Omega}_{t+1}) \\
        & d_{l'} = \textbf{argmax}(\boldsymbol{p}_{l'}) \\
        & \boldsymbol{D}_{l'+1} = \textbf{concat}(\boldsymbol{D}_{l'}, d_{l'}) \\
        & l' = l' + 1 \\
        \textbf{until: } & d_{l'} = <eos> \\
        \textbf{ret: } & \boldsymbol{D}_{L'=l'}
    \end{split}
\end{equation}

After the iteration is completed, the detokenization process is used to convert the output $\boldsymbol{D}_{L'}$ back into a string.

In the training phase, parallel training is adopted, based on masked self-attention mechanism, to ensure that only tokens before $d_{l'}$ can be seen when inferring $d_{l'+1}$. Given an input sequence of $\boldsymbol{D}_{\text{train}}=[d_0, d_1, ..., d_{L'}]$, model output logits of $\boldsymbol{P}\in \mathbb{R}^{(L'+1) \times \text{hidden\_size}}$, and ground truth of $\boldsymbol{D}_{\text{groud\_truth}}=[d_1, d_2, ..., d_{L'}, <eos>]$, focal loss \cite{lin2017focal} is taken as the loss function, as shown in Eq. {\ref{eq_sdec_train}}.

\begin{equation}
    \label{eq_sdec_train}
    \begin{split}
        \boldsymbol{P} &= \textbf{DecoderBlocks}(\boldsymbol{D}_{\text{train}}, \boldsymbol{\Omega_{t+1}}) \\
        \boldsymbol{P}' &= \textbf{softmax}(\boldsymbol{P}) \\
        \textbf{Loss} &= \sum_{i=1}^{L'}-(1-p_{d_i})^\gamma \log (p_{d_i}), \gamma=2
    \end{split}
\end{equation}

\subsection{Sentence-level Large Language Models (SLLMs)}

Currently, almost all decoder-only LLMs consist of the following components: an embedding layer $\textbf{EMB}_{\text{llm}}$, $N$ decoder-only blocks $\textbf{DB}_{\text{llm}}$, and a fully connected layer $\textbf{FC}_{\text{llm}}$ for outputting logits. As shown in Eq. \ref{eq_llm}, given context tokens $\boldsymbol{D}_{\text{context}}=[d_0, d_1, ..., d_{l'}]$, the model generates the next token $d_{l'+1}$.

\begin{equation}
    \label{eq_llm}
    \begin{split}
        \boldsymbol{E}_{\text{context}} &= \textbf{EMB}_{\text{llm}}(\boldsymbol{D}_{\text{context}}) \\
        \boldsymbol{H}_{\text{llm}} &= \textbf{DB}_{\text{llm}}(\boldsymbol{E}_{\text{context}}) \\
        \boldsymbol{P}_{\text{llm}} &= (p_{\text{llm},i})_{(l'+1) \times \text{hidden\_size}} \\
        & = \textbf{FC}_{\text{llm}}(\boldsymbol{H}_{\text{llm}}) \\
        d_{l'+1} &= \textbf{argmax}(\boldsymbol{p}_{\text{llm},l'})
    \end{split}
\end{equation}

In this section, we introduce a unified grafting method to graft SentienceVAE onto the beginning and end of any LLM. This method transforms the LLMs into SLLMs, enabling it to operate effectively in the sentence-level embedding space.

{\bf Discard Embedding Layer:} in a SLLM, the $N$ decoder-only blocks no longer receives word-level tokens directly. Instead, it receives sentence embedding vectors encoded by the sentence encoder. Consequently, the traditional embedding layer in the LLM architecture is removed. Therefore, the model follows Eq. \ref{eq_sllm_forward}.

\begin{equation}
    \label{eq_sllm_forward}
    \begin{split}
        \boldsymbol{E}_{\text{sllm}} &=  [\boldsymbol{\Omega}_1, \boldsymbol{\Omega}_2, ..., \boldsymbol{\Omega}_t] \\
        \boldsymbol{H}_{\text{sllm}} &= (\boldsymbol{h}_{\text{sllm},i})_{t \times \text{hidden\_size}} \\ 
        &= \textbf{DB}_\text{llm}(\boldsymbol{E}_{\text{sllm}})
    \end{split}
\end{equation}

{\bf Termination judgment layer:} in previous LLMs, the final fully connected layer $\textbf{FC}_{\text{llm}}$ typically converts an $\boldsymbol{H}_{\text{llm}} \in \mathbb{R}^{t \times \text{hidden\_size}}$ to a probability vector (logits) $\boldsymbol{P}_{\text{llm}} \in \mathbb{R}^{t \times V}$, where $V$ represents the size of the vocabulary. This allows the model to determine the next token output using sampling algorithms, \textit{e.g.}, Greedy Search,  Beam Search \cite{graves2012sequence}, Random Sampling, Temperature Sampling \cite{hinton2015distilling}, Top-K Sampling \cite{fan2018hierarchical}, and Nucleus Sampling \cite{holtzman2019curious}. If the current token output is a special token like eos (end-of-sequence), the iteration terminates. Notably, in SLLMs, the output of $\textbf{DB}_{\text{llm}}$ is a sentence-level hidden state vector $\boldsymbol{H}_{\text{sllm}}$. Therefore, traditional token generation methods are not applicable. Instead, we use a new fully connected layer $\textbf{FC}_{\text{sllm}}$ called termination judgment layer that converts the $\boldsymbol{H}_{\text{sllm}}$ into a 2-dimensional boolean vector $\boldsymbol{B}_{\text{stop\_flag}}$, as shown in Eq. \ref{eq_sllm_stop_flag}. This vector helps determine whether the current sentence-level hidden state vector signifies the end of a sentence (stop flag) or needs further decoding by the sentence decoder. If the vector indicates an end flag, the iteration terminates. Otherwise, the sentence decoder processes the embedding to generate corresponding tokens, as Eq. \ref{eq_sllm_fc}.

\begin{equation}
    \label{eq_sllm_stop_flag}
    \begin{split}
        \boldsymbol{B}_{\text{stop\_flag}} &= (\boldsymbol{b_i})_{t \times 2} \\ 
        &= \textbf{FC}_{\text{sllm}}(\boldsymbol{H}_{\text{sllm}})
    \end{split}
\end{equation}

\begin{equation}
    \label{eq_sllm_fc}
    \boldsymbol{\Omega}_{t+1} = 
    \begin{cases}
        <eos> &, \textbf{argmax}(\boldsymbol{b_t}) = 1 \\
        \boldsymbol{h}_{\text{sllm}, t} &, \textbf{argmax}(\boldsymbol{b_t}) = 0
    \end{cases}
\end{equation}

During the training phase, calculate the focal loss for $\boldsymbol{B}_{\text{stop\_flag}}$ as part of the global loss.

{\bf Inferencing sentence by sentence: } in the SLLM architecture, we use a tiny SentenceVAE model to encode multiple tokens within a sentence into a single sentence embedding vector. This enables the LLM to conduct inference on a sentence-by-sentence basis rather than the traditional token-by-token method.

\section{Experiment}

To validate our method, we initially trained individual SentienceVAEs using self-supervised methods, demonstrating the capability of compressing multiple tokens from a sentence into a single vector via an encoder and reconstructing the original sequence through a decoder. Subsequently, we integrated these encoders and decoders at the endpoints of open-source LLMs, thereby enhancing LLMs into SLLMs that operate within a sentence embedding space. This modification not only improves perplexity (PPL) and enhances inference speed, but also reduces memory consumption. Observations of the loss curve indicate that SLLMs continue to adhere to the Scaling Law \cite{kaplan2020scaling}.

\subsection{Experimental Setting}

Our experiments utilized either a single 4-card RTX 4090 (24G) or a 4-card A100 (40G) (for SLLM-1.3B), employing data parallel distribution for training.

{\bf Base Models:} we employed the 125M, 350M, and 1.3B models from the OPT series \cite{zhang2022opt} as our base LLMs, exploring the extension to larger models via the Scaling Law.

{\bf Dataset:} the training dataset comprised the English subset (EN/WebText) of the Wanjuan-1.0 dataset \cite{he2023wanjuan}, with SentenceVAE samples including approximately 153.6M sentences (1.7B tokens) and SLLM samples encompassing about 6.4M paragraphs (5.6B tokens). A validation set of 1,000 random sentences or paragraphs ensured no overlap with the training data.


{\bf Hyperparameters:} for SentenceVAEs, we set the maximum tokens per sentence at 64, with a batch size of 128/card, a base learning rate of 1e-7/batch/card, and up to 300K iterations. SLLMs training mirrored these settings, with a maximum of 64 sentences per paragraph, a batch size of 1/card, a base learning rate of 1e-6/batch/card, and up to 1.6M iterations.

All experiments utilized the AdamW optimizer \cite{loshchilov2017decoupled} with AMP \cite{micikevicius2017mixed} enabled, a weight decay coefficient of 0.01, and a maximum gradient L2 norm of 1 for clipping. The initial 5K iterations followed a linear learning rate schedule, transitioning to a cosine annealing schedule for subsequent iterations, alongside an Exponential Moving Average (EMA) strategy.

{\bf Metrics:} PPL was employed as the evaluation metric, calculated for output logits $\boldsymbol{P}$ and ground truth $\boldsymbol{D}_{\text{groud\_truth}}$ as per the defined formula Eq. \ref{eq_ppl}.

\begin{equation}
    \label{eq_ppl}
    \textbf{PPL} = \exp{( \frac{1}{L'}\sum_{i=1}^{L'}-\log(p_{d_i}))}
\end{equation}

\subsection{Sentence-level Tokens}

We trained SentienceVAEs using a self-supervised approach to assess if sentence embeddings could effectively represent and reconstruct word-level tokens. Models were trained with hidden sizes corresponding to the dimensions of the OPT-125M, OPT-350M, and OPT-1.3B, with varying block layers. Performance was evaluated using cross-entropy loss and PPL on the validation set, confirming our method. The experimental results are presented in Table \ref{tab_ppl_svae}.

\begin{table}[!h]
  \begin{center}
  \caption{Metrics of SentenceVAEs on validation set.}
  \label{tab_ppl_svae}
  \renewcommand{\arraystretch}{1.5}
  \begin{tabular}{c|cccc}
  \hline
  Model & \makecell{Hidden\\Size} & \makecell{Hidden\\Layers} & {Loss}$\downarrow$ & {PPL}$\downarrow$ \\
  \hline
  SVAE-768-H1 & 768 & 1 & 1.339 & 3.605 \\
  SVAE-768-H2 & 768 & 2 & 1.019 & 2.588 \\ 
  SVAE-768-H4 & 768 & 4 & \bf{0.5598} & \bf{1.649} \\ 
  \hline
  SVAE-1024-H1 & 1024 & 1 & 0.9266 & 2.406 \\
  SVAE-1024-H2 & 1024 & 2 & 0.6610 & 1.845 \\
  SVAE-1024-H4 & 1024 & 4 & \bf{0.3704} & \bf{1.384} \\
  \hline
  SVAE-2048-H1 & 2048 & 1 & 0.5165 & 1.622 \\ 
  SVAE-2048-H2 & 2048 & 2 & 0.2845 & 1.292 \\ 
  SVAE-2048-H4 & 2048 & 4 & \bf{0.1270} & \bf{1.115} \\
  \hline
  \end{tabular}
  \end{center}
\end{table}

\begin{table*}[!ht]
  \begin{center}
  \caption{Test examples of SentenceVAEs. Text in blue indicates out-of-distribution samples, text in red highlights partial mismatches between output and input, and the \textcolor{red}{$_{\triangle}$} denotes missing output.}
  \label{tab_svae_demo}
  \renewcommand{\arraystretch}{1.5}
  \begin{tabular}{p{0.02\textwidth}|p{0.2\textwidth}|p{0.04\textwidth}|p{0.2\textwidth}:p{0.2\textwidth}:p{0.2\textwidth}}
    \hline 
    \multirow{2}{*}{ID} & \multirow{2}{*}{Input Sentence} & \multirow{2}{*}{Tokens} & \multicolumn{3}{c}{Output Sentence}\\ 
    & & & \multicolumn{1}{>{\centering\arraybackslash}p{0.2\textwidth}}{SVAE-768-H4} & \multicolumn{1}{>{\centering\arraybackslash}p{0.2\textwidth}}{SVAE-1024-H4} & \multicolumn{1}{>{\centering\arraybackslash}p{0.2\textwidth}}{SVAE-2048-H4} \\
    \hline 
    1 & Hello, & 3 & Hello, & Hello, & Hello, \\
    \hdashline
    2 & What's your name? & 6 & What's your name? & What's your name? & What's your name? \\ 
    \hdashline 
    3 & What's your problem? & 6 & What's your problem? & What's your problem? & What's your problem? \\
    \hdashline
    4 & Hello, \textcolor{blue}{my dear friend} & 6 & Hello \textcolor{red}{$_{\triangle}$} my dear friend\textcolor{red}{,} & Hello \textcolor{red}{$_{\triangle}$} my dear friend\textcolor{red}{,} & Hello \textcolor{red}{$_{\triangle}$} my dear friend\textcolor{red}{,} \\
    \hdashline
    5 & Today is Friday. & 5 & Today is Friday. & Today is Friday. & Today is Friday. \\
    \hdashline
    6 & One two three four five six seven eight nine ten\~ & 12 & One two three four \textcolor{red}{six nine ten} eight nine \textcolor{red}{night} & One two three \textcolor{red}{six} four \textcolor{red}{three} seven eight nine ten\~ & One two three four five six \textcolor{red}{eight seven} nine\textcolor{red}{\~ \ } ten \textcolor{red}{$_{\triangle}$} \\
    \hdashline
    7 & Hahaha.\textcolor{blue}{. and you?} & 8 & Hahaha\textcolor{red}{aha} and you? & Hahaha\textcolor{red}{aha} and you? & Hahaha\textcolor{red}{aha} and you? \\ 
    \hdashline
    8 & \textcolor{blue}{Yao yao ling xian!} & 9 & Yao\textcolor{red}{aoaoao} xian! & yao \textcolor{red}{yaoaoao} xian! & Yao yao ling xian! \\
    \hline
  \end{tabular}
  \end{center}
\end{table*}

\begin{figure*}[!ht]
    \centering
    \includegraphics[clip, trim=5pt 5pt 20pt 5pt, width=2.0\columnwidth]{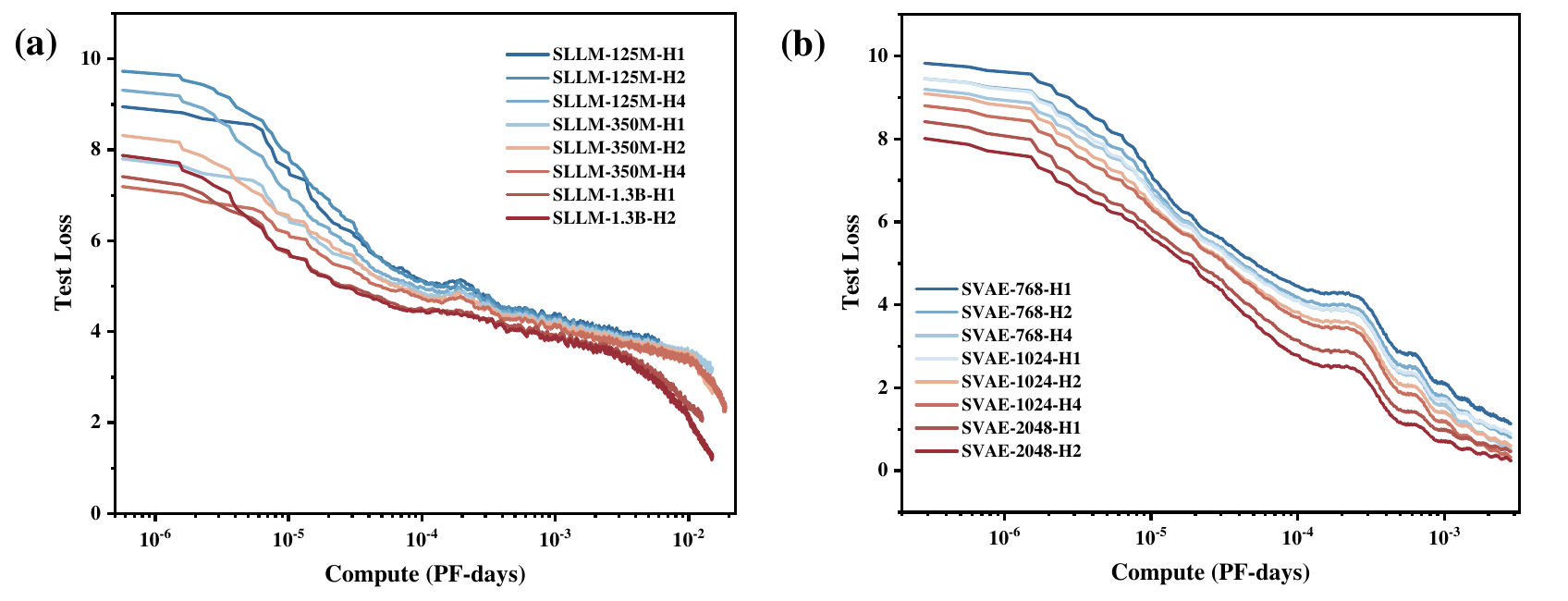}
  \caption{Scaling Law of (a) SLLMs and (b) SVAEs.}
  \label{fig:scaling}
\end{figure*}

The evaluation results confirm that SentenceVAEs function as intended,validating their theoretical foundations. Table \ref{tab_svae_demo} showcases a variety of test cases, including \textbf{Samples 1, 2, 3, 5, 6}, which serve as conventional examples. To evaluate model robustness, several out-of-distribution samples, marked in blue, are also incorporated:

\begin{itemize}
    \item {\bf Sample 4, 7:} data preprocessing entails segmenting sentences based on punctuation marks, such as \texttt{","} and \texttt{"."}. Consequently, these samples are split into multiple inputs during both training and deployment. Inputting these samples without prior preprocessing constitutes invalid input for the model.
    \item {\bf Sample 8:} this sample is the pinyin transcription of a well-known Chinese saying and is not in English, hence it is considered invalid input.
\end{itemize}

Analysis of the model's output reveals the following:

\begin{itemize}
    \item In the conventional samples, the model accurately reconstructed the sentences, with the exception of \textbf{Sample 6}. The error in \textbf{Sample 6} may be attributed to limitations inherent in the sinusoidal position encoding mechanism \cite{vaswani2017attention}.
   \item Although \textbf{Sample 4} contains the illegal input, the model has accurately restored it with only error of punctuation marks. \textbf{Sample 7} still maintains the meaning of the sentence. In \textbf{Sample 8}, as the hidden size of the model increases, the model also accurately restores the illegal input ``yao yao ling xian''. These phenomena indicate that the model is robust to illegal inputs.
\end{itemize}

Overall, the experimental results indicate that, a series of word-level tokens can be represented by a sentence-level token with strong robustness.

\subsection{Sentence-level LLMs (SLLMs)}

Upon validating the practicality of sentence-level embeddings, we integrated the encoder and decoder components from SentenceVAE with the OPT series models. This integration is detailed in Table \ref{tab_info_sllm}.

\begin{table}[!ht]
    \begin{center}
  \caption{The correspondence relationship between SLLMs, base LLMs, and SentenceVAEs}
  \label{tab_info_sllm}
  \begin{tabular}{c|cc}
  \hline
  Model & \makecell{Base\\Model} & \makecell{Grafted\\SentenceVAE} \\
  \hline 
  SLLM-125M-H1 & \multirow{3}{*}{OPT-125M} & SVAE-768-H1 \\
  SLLM-125M-H2 & & SVAE-768-H2 \\ 
  SLLM-125M-H4 & & SVAE-768-H4 \\
  \hline 
  SLLM-350M-H1 & \multirow{3}{*}{OPT-350M} & SVAE-1024-H1 \\ 
  SLLM-350M-H2 & & SVAE-1024-H2 \\
  SLLM-350M-H4 & & SVAE-1024-H4 \\
  \hline 
  SLLM-1.3B-H1 & \multirow{2}{*}{OPT-1.3B} & SVAE-2048-H1 \\ 
  SLLM-1.3B-H2 & & SVAE-2048-H2 \\
  \hline
  \end{tabular}
  \end{center}
\end{table}

\begin{table*}[!h]
    \begin{center}
  \caption{Benchmark test results of OPTs and SLLMs.}
  \label{tab_speed_sllm}
  \begin{tabular}{cc|ccc:ccc:ccc}
  \hline
  \multirow{2}{*}{Model} & \multirow{2}{*}{\makecell{Total\\Params}} & \multicolumn{3}{c}{\makecell{Average PPL}} & \multicolumn{3}{c}{\makecell{Mean output throughput\\(toks/s)}} & \multicolumn{3}{c}{\makecell{Mean GPU memory\\(KB/token)}} \\
  & & OPT$\downarrow$ & \textbf{SLLM}$\downarrow$ & $\Delta \downarrow$ & OPT$\uparrow$ & \textbf{SLLM}$\uparrow$ & $\Delta \uparrow$ & OPT$\downarrow$ & \textbf{SLLM}$\downarrow$ & $\Delta \downarrow$ \\
  \hline 
  SLLM-125M-H1 & 214M & \multirow{3}{*}{26.75} & 31.68 & +18.4\% & \multirow{3}{*}{214.57} & \textbf{652.78} & \textbf{+204.2\%} & \multirow{3}{*}{73.15} & 12.03 & -83.6\% \\ 
  SLLM-125M-H2 & 226M & & 44.60 & +66.7\% &  & 539.80 & +151.6\% & & \textbf{7.08} & \textbf{-90.3\%} \\
  SLLM-125M-H4 & 250M & & \textbf{14.32} & \textbf{-46.5\%} &  & 332.12 & +54.8\% & & 10.00 & -86.3\% \\
  \hline 
  SLLM-350M-H1 & 429M & \multirow{3}{*}{25.18} & 24.84 & -1.4\% & \multirow{3}{*}{144.33} & \textbf{481.39} & \textbf{+233.5\%} & \multirow{3}{*}{197.59} & 29.98 & -84.8\% \\
  SLLM-350M-H2 & 450M &  & 14.81 & -41.2\% & & 442.23 & +206.4\% & & 26.78 & -86.4\% \\
  SLLM-350M-H4 & 492M &  & \textbf{10.17} & \textbf{-59.6\%} & & 315.61 & +118.7\% & & \textbf{17.73} & \textbf{-91.0\%} \\
  \hline 
  SLLM-1.3B-H1 & 1.61B &  \multirow{2}{*}{15.95} & 8.76 & -45.1\% & \multirow{2}{*}{119.07} & 479.71 & +302.9\% & \multirow{2}{*}{400.01} & 57.07 & -85.7\% \\
  SLLM-1.3B-H2 & 1.69B &  & \textbf{3.84} & \textbf{-75.9\%} & & \textbf{553.95} & \textbf{+365.2\%} & & \textbf{55.14} & \textbf{-86.2\%} \\
  \hline
  \end{tabular}
  \end{center}
\end{table*}

Due to memory constraints during data-parallel distributed training, we were unable to evaluate the SLLM-1.3B-H4 model. Comparative performance metrics for the OPT and SLLM models are presented in Table \ref{tab_speed_sllm}.

\textbf{Faster Inference:} the throughput of SLLMs, measured using the PyTorch framework \cite{imambi2021pytorch}, was assessed without the inclusion of optimization techniques such as PagedAttention \cite{kwon2023efficient} or acceleration frameworks like TensorRT-LLM or LMDeploy \cite{2023lmdeploy}. The baseline for this comparison was the original OPT model's throughput. Experimental findings indicate that SLLMs achieve an average inference speed that is 2 to 3 times faster than traditional LLMs. Notably, as the model size increases, the relative overhead of SentenceVAE decreases, thereby magnifying the speed advantage.

\textbf{Higher Accuracy:} the PPL scores for SLLMs were superior to those of the baseline OPT models. This enhancement is attributed to the SLLM framework's ability to process language at a more granular sentence level, thereby boosting overall performance. Moreover, the computation of attention within the sentence embedding space enables SLLMs to handle shorter relative contexts for equivalent text lengths, further enhancing model efficacy.

\textbf{Longer Context:} The maximum context length a model can handle is typically constrained by GPU memory availability. The SLLM framework addresses this by compressing multiple original tokens into a single token, reducing memory usage for contexts of equivalent length. This compression enables an extended context capacity within the same hardware limitations.

In summary, SLLMs can work in setence-level embedding space with faster inference speed, more accurate PPL, and longer context.

\subsection{Scaling Law of SLLMs}

During the training of SLLMs, an analysis of the loss curve indicates compliance with the Scaling Law, as depicted in Figure \ref{fig:scaling}. This result indicates that SLLMs also adhere to the Scaling Law, suggesting that this theory can be extended to larger models.

\section{Conclusion \& Future Trends}

This section encapsulates the contributions of our research and delineates several promising avenues for exploration.

\subsection{Scaling up SLLMs with Enhanced Architectures}

In our study, we employed the well-established Transformer architecture \cite{vaswani2017attention}, implemented using the PyTorch framework \cite{imambi2021pytorch}, to expedite the validation of our hypotheses. Constraints related to computational resources and time necessitated that our evaluations were confined to models with parameter sizes ranging from 125M to 1.3B. Nevertheless, by corroborating the Scaling Law, we extrapolated the feasibility of our methodologies to larger-scale models. Future iterations of the SLLM framework could integrate advanced architectural enhancements, such as Rotational Position Encoding (RoPE) \cite{su2024roformer}, to rectify sequence ordering issues identified in \textbf{Sample 6}. Although our current models are trained exclusively on English-language corpora, extending support to multiple languages could be a beneficial direction. Despite its nascent stage, the SLLM framework exhibits substantial potential for enhancements and broader applicability.

\subsection{SLLMs in Hybrid Edge-Cloud Inference}

The SLLM framework embodies a hybrid architecture that amalgamates small and large models. By pinpointing an optimal balance, the smaller model (SentenceVAE) and the larger model (LLM) can be efficiently allocated across edge and cloud environments, optimizing computational loads and improving user interactions.

\subsection{SLLMs and Embodied Intelligence}

Currently, embodied intelligence that leverages the LLM Agent paradigm faces challenges in direct hardware interactions, relying instead on intermediary mechanisms to translate high-level LLM directives into conventional real-time control logic. A critical bottleneck is the generation rate of ``tokens'' at the edge. In contrast, the SLLM framework's enhanced ability to process and produce more ``tokens'' under identical computational and temporal constraints presents opportunities for direct interfacing of large embodied intelligence models with underlying hardware systems.

\subsection{SLLMs in Multimodal Large Models}

In the context of multimodal large models, the ``frames'' in video and ``trunks'' in audio data can be analogized to ``tokens'' in text-based models. The SLLM method could thus be adapted to augment the processing rates of ``frames'' in these modalities, potentially elevating user experiences and achieving performance that matches or surpasses current standards.

\section{Acknowledgments}

This research was supported by the China National Key R\&D Program (2022YFC2808003) and the Natural Science Basic Research Program of Shaanxi (2024JC-YBMS-468).

\bibliography{aaai25}

\begin{thebibliography}{38}
\providecommand{\natexlab}[1]{#1}

\bibitem[{Ba, Kiros, and Hinton(2016)}]{ba2016layer}
Ba, J.~L.; Kiros, J.~R.; and Hinton, G.~E. 2016.
\newblock Layer Normalization.
\newblock \emph{arXiv preprint arXiv:1607.06450}.

\bibitem[{Badrinarayanan, Kendall, and Cipolla(2017)}]{badrinarayanan2017segnet}
Badrinarayanan, V.; Kendall, A.; and Cipolla, R. 2017.
\newblock Segnet: A deep convolutional encoder-decoder architecture for image segmentation.
\newblock \emph{IEEE transactions on pattern analysis and machine intelligence}, 39(12): 2481--2495.

\bibitem[{Cai et~al.(2024)Cai, Cao, Chen, Chen, Chen, Chen, Chen, Chen, Chen, Chu, Dong, Duan, Fan, Fei, Gao, Ge, Gu, Gu, Gui, Guo, Guo, He, Hu, Huang, Jiang, Jiao, Jin, Lei, Li, Li, Li, Li, Li, Li, Liu, Liu, Hong, Liu, Liu, Liu, Lv, Lv, Lv, Ma, Ma, Ma, Ning, Ouyang, Qiu, Qu, Shang, Shao, Song, Song, Sui, Sun, Sun, Tang, Wang, Wang, Wang, Wang, Wang, Wang, Wang, Wei, Weng, Wu, Xiong, Xu, Xu, Yan, Yan, Yang, Ye, Ying, Yu, Yu, Zang, Zhang, Zhang, Zhang, Zhang, Zhang, Zhang, Zhang, Zhang, Zhang, Zhang, Zhang, Zhao, Zhao, Zhao, Zhou, Zhou, Zhuo, Zou, Qiu, Qiao, and Lin}]{cai2024internlm2}
Cai, Z.; Cao, M.; Chen, H.; Chen, K.; Chen, K.; Chen, X.; Chen, X.; Chen, Z.; Chen, Z.; Chu, P.; Dong, X.; Duan, H.; Fan, Q.; Fei, Z.; Gao, Y.; Ge, J.; Gu, C.; Gu, Y.; Gui, T.; Guo, A.; Guo, Q.; He, C.; Hu, Y.; Huang, T.; Jiang, T.; Jiao, P.; Jin, Z.; Lei, Z.; Li, J.; Li, J.; Li, L.; Li, S.; Li, W.; Li, Y.; Liu, H.; Liu, J.; Hong, J.; Liu, K.; Liu, K.; Liu, X.; Lv, C.; Lv, H.; Lv, K.; Ma, L.; Ma, R.; Ma, Z.; Ning, W.; Ouyang, L.; Qiu, J.; Qu, Y.; Shang, F.; Shao, Y.; Song, D.; Song, Z.; Sui, Z.; Sun, P.; Sun, Y.; Tang, H.; Wang, B.; Wang, G.; Wang, J.; Wang, J.; Wang, R.; Wang, Y.; Wang, Z.; Wei, X.; Weng, Q.; Wu, F.; Xiong, Y.; Xu, C.; Xu, R.; Yan, H.; Yan, Y.; Yang, X.; Ye, H.; Ying, H.; Yu, J.; Yu, J.; Zang, Y.; Zhang, C.; Zhang, L.; Zhang, P.; Zhang, P.; Zhang, R.; Zhang, S.; Zhang, S.; Zhang, W.; Zhang, W.; Zhang, X.; Zhang, X.; Zhao, H.; Zhao, Q.; Zhao, X.; Zhou, F.; Zhou, Z.; Zhuo, J.; Zou, Y.; Qiu, X.; Qiao, Y.; and Lin, D. 2024.
\newblock InternLM2 Technical Report.
\newblock arXiv:2403.17297.

\bibitem[{Chen et~al.(2023)Chen, Sun, Li, and Li}]{chen2023computational}
Chen, Y.; Sun, Z.; Li, C.; and Li, X. 2023.
\newblock Computational ghost imaging in turbulent water based on self-supervised information extraction network.
\newblock \emph{Optics \& Laser Technology}, 167: 109735.

\bibitem[{Cho et~al.(2014)Cho, Van~Merri{\"e}nboer, Bahdanau, and Bengio}]{cho2014properties}
Cho, K.; Van~Merri{\"e}nboer, B.; Bahdanau, D.; and Bengio, Y. 2014.
\newblock On the properties of neural machine translation: Encoder-decoder approaches.
\newblock \emph{arXiv preprint arXiv:1409.1259}.

\bibitem[{Contributors(2023)}]{2023lmdeploy}
Contributors, L. 2023.
\newblock LMDeploy: A Toolkit for Compressing, Deploying, and Serving LLM.
\newblock \url{https://github.com/InternLM/lmdeploy}.

\bibitem[{Devlin et~al.(2018)Devlin, Chang, Lee, and Toutanova}]{devlin2018bert}
Devlin, J.; Chang, M.-W.; Lee, K.; and Toutanova, K. 2018.
\newblock Bert: Pre-training of deep bidirectional transformers for language understanding.
\newblock \emph{arXiv preprint arXiv:1810.04805}.

\bibitem[{Fan, Lewis, and Dauphin(2018)}]{fan2018hierarchical}
Fan, A.; Lewis, M.; and Dauphin, Y. 2018.
\newblock Hierarchical Neural Story Generation.
\newblock \emph{arXiv preprint arXiv:1805.04833}.

\bibitem[{Gloeckle et~al.(2024)Gloeckle, Idrissi, Rozi{\`e}re, Lopez-Paz, and Synnaeve}]{gloeckle2024better}
Gloeckle, F.; Idrissi, B.~Y.; Rozi{\`e}re, B.; Lopez-Paz, D.; and Synnaeve, G. 2024.
\newblock Better \& Faster Large Language Models via Multi-token Prediction.
\newblock \emph{arXiv preprint arXiv:2404.19737}.

\bibitem[{Graves(2012)}]{graves2012sequence}
Graves, A. 2012.
\newblock Sequence Transduction with Recurrent Neural Networks.
\newblock \emph{arXiv preprint arXiv:1211.3711}.

\bibitem[{He et~al.(2023)He, Jin, Xu, Qiu, Wang, Li, Yan, Wang, and Lin}]{he2023wanjuan}
He, C.; Jin, Z.; Xu, C.; Qiu, J.; Wang, B.; Li, W.; Yan, H.; Wang, J.; and Lin, D. 2023.
\newblock WanJuan: A Comprehensive Multimodal Dataset for Advancing English and Chinese Large Models.
\newblock arXiv:2308.10755.

\bibitem[{Hinton, Vinyals, and Dean(2015)}]{hinton2015distilling}
Hinton, G.; Vinyals, O.; and Dean, J. 2015.
\newblock Distilling the Knowledge in a Neural Network.
\newblock \emph{arXiv preprint arXiv:1503.02531}.

\bibitem[{Holtzman et~al.(2019)Holtzman, Buys, Du, Forbes, and Choi}]{holtzman2019curious}
Holtzman, A.; Buys, J.; Du, L.; Forbes, M.; and Choi, Y. 2019.
\newblock The Curious Case of Neural Text Degeneration.
\newblock \emph{arXiv preprint arXiv:1904.09751}.

\bibitem[{Imambi, Prakash, and Kanagachidambaresan(2021)}]{imambi2021pytorch}
Imambi, S.; Prakash, K.~B.; and Kanagachidambaresan, G. 2021.
\newblock PyTorch.
\newblock \emph{Programming with TensorFlow: solution for edge computing applications}, 87--104.

\bibitem[{Islam and Moushi(2024)}]{islam2024gpt}
Islam, R.; and Moushi, O.~M. 2024.
\newblock GPT-4o: The Cutting-Edge Advancement in Multimodal LLM.
\newblock \emph{Authorea Preprints}.

\bibitem[{Kaplan et~al.(2020)Kaplan, McCandlish, Henighan, Brown, Chess, Child, Gray, Radford, Wu, and Amodei}]{kaplan2020scaling}
Kaplan, J.; McCandlish, S.; Henighan, T.; Brown, T.~B.; Chess, B.; Child, R.; Gray, S.; Radford, A.; Wu, J.; and Amodei, D. 2020.
\newblock Scaling Laws for Neural Language Models.
\newblock \emph{arXiv preprint arXiv:2001.08361}.

\bibitem[{Kingma and Welling(2013)}]{kingma2013auto}
Kingma, D.~P.; and Welling, M. 2013.
\newblock Auto-encoding variational bayes.
\newblock \emph{arXiv preprint arXiv:1312.6114}.

\bibitem[{Kwon et~al.(2023)Kwon, Li, Zhuang, Sheng, Zheng, Yu, Gonzalez, Zhang, and Stoica}]{kwon2023efficient}
Kwon, W.; Li, Z.; Zhuang, S.; Sheng, Y.; Zheng, L.; Yu, C.~H.; Gonzalez, J.; Zhang, H.; and Stoica, I. 2023.
\newblock Efficient Memory Management for Large Language Model Serving with PagedAttention.
\newblock In \emph{Proceedings of the 29th Symposium on Operating Systems Principles}, 611--626.

\bibitem[{Lazaridou et~al.(2022)Lazaridou, Gribovskaya, Stokowiec, and Grigorev}]{lazaridou2022internet}
Lazaridou, A.; Gribovskaya, E.; Stokowiec, W.; and Grigorev, N. 2022.
\newblock Internet-augmented language models through few-shot prompting for open-domain question answering.
\newblock \emph{arXiv preprint arXiv:2203.05115}.

\bibitem[{Le~Cun(1987)}]{lecun1987}
Le~Cun, Y. 1987.
\newblock \emph{Modèles connexionnistes de l'apprentissage}.
\newblock Ph.D. thesis, Université Pierre et Marie Curie.
\newblock Thèse de doctorat dirigée par Milgram, Maurice Sciences appliquées Paris 6 1987.

\bibitem[{Li et~al.(2020)Li, Liu, Wu, Liu, Zhao, and Liu}]{li2020robutrans}
Li, N.; Liu, Y.; Wu, Y.; Liu, S.; Zhao, S.; and Liu, M. 2020.
\newblock Robutrans: A robust transformer-based text-to-speech model.
\newblock In \emph{Proceedings of the AAAI conference on artificial intelligence}, volume~34, 8228--8235.

\bibitem[{Li et~al.(2024{\natexlab{a}})Li, Chen, Tian, and Sun}]{li2024part}
Li, X.; Chen, Y.; Tian, T.; and Sun, Z. 2024{\natexlab{a}}.
\newblock Part-based image-loop network for single-pixel imaging.
\newblock \emph{Optics \& Laser Technology}, 168: 109917.

\bibitem[{Li et~al.(2024{\natexlab{b}})Li, Yao, Jiang, Fang, Wang, Liu, Wang, Zhao, Wang, Huang, Song, Li, Zhang, Zhao, Sun, Wang, He, Wang, Li, and Huang}]{li2024teleflm}
Li, X.; Yao, Y.; Jiang, X.; Fang, X.; Wang, C.; Liu, X.; Wang, Z.; Zhao, Y.; Wang, X.; Huang, Y.; Song, S.; Li, Y.; Zhang, Z.; Zhao, B.; Sun, A.; Wang, Y.; He, Z.; Wang, Z.; Li, X.; and Huang, T. 2024{\natexlab{b}}.
\newblock Tele-FLM Technical Report.
\newblock arXiv:2404.16645.

\bibitem[{Lin et~al.(2017)Lin, Goyal, Girshick, He, and Doll{\'a}r}]{lin2017focal}
Lin, T.-Y.; Goyal, P.; Girshick, R.; He, K.; and Doll{\'a}r, P. 2017.
\newblock Focal Loss for Dense Object Detection.
\newblock In \emph{Proceedings of the IEEE international conference on computer vision}, 2980--2988.

\bibitem[{Loshchilov and Hutter(2017)}]{loshchilov2017decoupled}
Loshchilov, I.; and Hutter, F. 2017.
\newblock Decoupled Weight Decay Regularization.
\newblock \emph{arXiv preprint arXiv:1711.05101}.

\bibitem[{Makin, Moses, and Chang(2020)}]{makin2020machine}
Makin, J.~G.; Moses, D.~A.; and Chang, E.~F. 2020.
\newblock Machine translation of cortical activity to text with an encoder--decoder framework.
\newblock \emph{Nature neuroscience}, 23(4): 575--582.

\bibitem[{Micikevicius et~al.(2017)Micikevicius, Narang, Alben, Diamos, Elsen, Garcia, Ginsburg, Houston, Kuchaiev, Venkatesh et~al.}]{micikevicius2017mixed}
Micikevicius, P.; Narang, S.; Alben, J.; Diamos, G.; Elsen, E.; Garcia, D.; Ginsburg, B.; Houston, M.; Kuchaiev, O.; Venkatesh, G.; et~al. 2017.
\newblock Mixed Precision Training.
\newblock \emph{arXiv preprint arXiv:1710.03740}.

\bibitem[{Raffel et~al.(2020)Raffel, Shazeer, Roberts, Lee, Narang, Matena, Zhou, Li, and Liu}]{raffel2020exploring}
Raffel, C.; Shazeer, N.; Roberts, A.; Lee, K.; Narang, S.; Matena, M.; Zhou, Y.; Li, W.; and Liu, P.~J. 2020.
\newblock Exploring the limits of transfer learning with a unified text-to-text transformer.
\newblock \emph{Journal of machine learning research}, 21(140): 1--67.

\bibitem[{Singhal et~al.(2023)Singhal, Tu, Gottweis, Sayres, Wulczyn, Hou, Clark, Pfohl, Cole-Lewis, Neal et~al.}]{singhal2023towards}
Singhal, K.; Tu, T.; Gottweis, J.; Sayres, R.; Wulczyn, E.; Hou, L.; Clark, K.; Pfohl, S.; Cole-Lewis, H.; Neal, D.; et~al. 2023.
\newblock Towards expert-level medical question answering with large language models.
\newblock \emph{arXiv preprint arXiv:2305.09617}.

\bibitem[{Su et~al.(2024)Su, Ahmed, Lu, Pan, Bo, and Liu}]{su2024roformer}
Su, J.; Ahmed, M.; Lu, Y.; Pan, S.; Bo, W.; and Liu, Y. 2024.
\newblock RoFormer: Enhanced Transformer with Rotary Position Embedding.
\newblock \emph{Neurocomputing}, 568: 127063.

\bibitem[{Sutskever, Vinyals, and Le(2014)}]{sutskever2014sequence}
Sutskever, I.; Vinyals, O.; and Le, Q.~V. 2014.
\newblock Sequence to sequence learning with neural networks.
\newblock \emph{Advances in neural information processing systems}, 27.

\bibitem[{Touvron et~al.(2023{\natexlab{a}})Touvron, Lavril, Izacard, Martinet, Lachaux, Lacroix, Rozi{\`e}re, Goyal, Hambro, Azhar et~al.}]{touvron2023llama}
Touvron, H.; Lavril, T.; Izacard, G.; Martinet, X.; Lachaux, M.-A.; Lacroix, T.; Rozi{\`e}re, B.; Goyal, N.; Hambro, E.; Azhar, F.; et~al. 2023{\natexlab{a}}.
\newblock LLaMA: Open and Efficient Foundation Language Models.
\newblock \emph{arXiv preprint arXiv:2302.13971}.

\bibitem[{Touvron et~al.(2023{\natexlab{b}})Touvron, Martin, Stone, Albert, Almahairi, Babaei, Bashlykov, Batra, Bhargava, Bhosale et~al.}]{touvron2023llama2}
Touvron, H.; Martin, L.; Stone, K.; Albert, P.; Almahairi, A.; Babaei, Y.; Bashlykov, N.; Batra, S.; Bhargava, P.; Bhosale, S.; et~al. 2023{\natexlab{b}}.
\newblock Llama 2: Open Foundation and Fine-Tuned Chat Models.
\newblock \emph{arXiv preprint arXiv:2307.09288}.

\bibitem[{Vaswani et~al.(2017)Vaswani, Shazeer, Parmar, Uszkoreit, Jones, Gomez, Kaiser, and Polosukhin}]{vaswani2017attention}
Vaswani, A.; Shazeer, N.; Parmar, N.; Uszkoreit, J.; Jones, L.; Gomez, A.~N.; Kaiser, {\L}.; and Polosukhin, I. 2017.
\newblock Attention is All you Need.
\newblock \emph{Advances in neural information processing systems}, 30.

\bibitem[{Wang et~al.(2023)Wang, Lyu, Ji, Zhang, Yu, Shi, and Tu}]{wang-etal-2023-document-level}
Wang, L.; Lyu, C.; Ji, T.; Zhang, Z.; Yu, D.; Shi, S.; and Tu, Z. 2023.
\newblock Document-Level Machine Translation with Large Language Models.
\newblock In Bouamor, H.; Pino, J.; and Bali, K., eds., \emph{Proceedings of the 2023 Conference on Empirical Methods in Natural Language Processing}, 16646--16661. Singapore: Association for Computational Linguistics.

\bibitem[{Wang et~al.(2024)Wang, Liu, Liu, Yao, Huang, He, Li, Li, Che, Zhang, Wang, Wang, Pu, Xu, Fang, Zhao, Zhang, Huang, Lu, Peng, Zheng, Wang, Yang, he, Jiang, Xie, Zhang, Li, Shi, Fu, Zhang, Huang, Xiong, Zhang, Wang, and Song}]{wang2024telechat}
Wang, Z.; Liu, X.; Liu, S.; Yao, Y.; Huang, Y.; He, Z.; Li, X.; Li, Y.; Che, Z.; Zhang, Z.; Wang, Y.; Wang, X.; Pu, L.; Xu, H.; Fang, R.; Zhao, Y.; Zhang, J.; Huang, X.; Lu, Z.; Peng, J.; Zheng, W.; Wang, S.; Yang, B.; he, X.; Jiang, Z.; Xie, Q.; Zhang, Y.; Li, Z.; Shi, L.; Fu, W.; Zhang, Y.; Huang, Z.; Xiong, S.; Zhang, Y.; Wang, C.; and Song, S. 2024.
\newblock TeleChat Technical Report.
\newblock arXiv:2401.03804.

\bibitem[{Zhang, Haddow, and Birch(2023)}]{pmlr-v202-zhang23m}
Zhang, B.; Haddow, B.; and Birch, A. 2023.
\newblock Prompting Large Language Model for Machine Translation: A Case Study.
\newblock In Krause, A.; Brunskill, E.; Cho, K.; Engelhardt, B.; Sabato, S.; and Scarlett, J., eds., \emph{Proceedings of the 40th International Conference on Machine Learning}, volume 202 of \emph{Proceedings of Machine Learning Research}, 41092--41110. PMLR.

\bibitem[{Zhang et~al.(2022)Zhang, Roller, Goyal, Artetxe, Chen, Chen, Dewan, Diab, Li, Lin et~al.}]{zhang2022opt}
Zhang, S.; Roller, S.; Goyal, N.; Artetxe, M.; Chen, M.; Chen, S.; Dewan, C.; Diab, M.; Li, X.; Lin, X.~V.; et~al. 2022.
\newblock OPT: Open Pre-trained Transformer Language Models.
\newblock \emph{arXiv preprint arXiv:2205.01068}.

\end{thebibliography}

\end{document}